%% file: paper.tex
\documentclass[]{fairmeta}

\usepackage{fvextra}
\DefineVerbatimEnvironment{VerbatimWrap}{Verbatim}{
breaklines=true, 
breakindent=0pt, 
breaksymbol={},
fontsize=\small,
}
\usepackage{xspace}
\usepackage{amssymb}%
\usepackage{pifont}%
\title{FACTORY: A Challenging Human-Verified Prompt Set for Long-Form Factuality}

\author[]{Mingda Chen}
\author[]{Yang Li}
\author[]{Xilun Chen}
\author[]{Adina Williams}
\author[]{Gargi Ghosh}
\author[]{Scott Yih}

\affiliation[]{FAIR at Meta}

\abstract{Long-form factuality evaluation assesses the ability of models to generate accurate, comprehensive responses to short prompts.
Existing benchmarks often lack human verification, leading to potential quality issues. To address this limitation, we introduce \ourdataset, a large-scale, human-verified prompt set. Developed using a model-in-the-loop approach and refined by humans, \ourdataset includes challenging prompts that are fact-seeking, answerable, and unambiguous.
We conduct human evaluations on 6 state-of-the-art language models using \ourdataset and existing datasets. Our results show that \ourdataset is a challenging benchmark: approximately 40\% of the claims made in the responses of SOTA models are not factual, compared to only 10\% for other datasets.
Our analysis identifies the strengths of \ourdataset over prior benchmarks, emphasizing its reliability and the necessity for models to reason across long-tailed facts.}

\date{\today}
\correspondence{Mingda Chen at \email{mingdachen@meta.com} Scott Yih at \email{scottyih@meta.com}}

\metadata[Dataset]{\url{https://huggingface.co/datasets/facebook/FACTORY}}

\newcommand{\cmark}{\ding{51}}%
\newcommand{\xmark}{\ding{55}}%

\newcommand{\ourdataset}{FACTORY\xspace}

\newcommand{\vs}{VeriScore\xspace}
\newcommand{\longfact}{LongFact\xspace}
\newcommand{\factbench}{FactBench\xspace}

\newcommand{\gpt}{GPT-4o\xspace}
\newcommand{\deepseek}{DeepSeek V3\xspace}
\newcommand{\gemini}{Gemini 2.5 Pro\xspace}
\newcommand{\claude}{Claude 3.7 Sonnet\xspace}

\newcommand{\llama}{Llama 4 Maverick\xspace}
\newcommand{\mds}{MassiveDS\xspace}

\begin{document}

\maketitle
\input{sections/intro}

\input{sections/related}

\input{sections/factory}

\input{sections/experiments}

\input{sections/conclusion}

\clearpage
\newpage
\bibliographystyle{assets/plainnat}
\bibliography{paper,anthology}

\clearpage
\newpage
\beginappendix

\input{sections/appendix}

\end{document}

%% file: sections/intro.tex
\section{Introduction}

Long-form factuality aims to assess models' abilities to produce comprehensive and accurate responses to relatively short prompts.
Existing benchmark prompt sets are typically generated automatically, often missing crucial human verification for quality assurance.
While considerable research has been devoted to improving models' factuality~\citep{zhang-etal-2024-self,xie2024improving,chen2024improving}, there is a notable gap in developing reliable benchmarks.

To bridge this gap, we introduce \ourdataset\footnote{\textbf{F}ramework for \textbf{A}ssessing \textbf{C}ontextual \textbf{T}ruth in \textbf{O}pen Book Long-Fo\textbf{r}m Factualit\textbf{y}}, a large-scale, human-verified, and challenging prompt set.
We employ a model-in-the-loop approach to ensure quality and address the complexities of evaluating long-form generation. Starting with seed topics from Wikipedia, we expand each topic into a diverse set of prompts using large language models (LLMs). We then apply the model-in-the-loop method to filter out simpler prompts, maintaining a high level of difficulty. Human annotators further refine the prompts to ensure they are fact-seeking, answerable, unambiguous, not time-sensitive, and safe. To push the boundaries of long-form factuality evaluation, we identify a ``hard'' split of \ourdataset that presents significant challenges to current state-of-the-art LLMs, with their outputs containing approximately 40\% of claims for which humans cannot find supportive information online.

In our experiments, we benchmark 6 state-of-the-art LLMs on \ourdataset and two existing benchmarks. The human evaluations consistently show that state-of-the-art LLMs achieve around 90\% factual precision on existing benchmarks. However, \ourdataset proves more challenging, as the factual precision of the SOTA LLMs are only approximately 60\% on its hard subset. These findings indicate that \ourdataset presents genuine challenges to current state-of-the-art LLMs.

We then conduct a detailed analysis to identify the characteristics of \ourdataset and the quality issues in prior benchmarks.
We discover that prior benchmarks suffer from issues such as answerability, hallucinations, and time-sensitivity, likely due to their lack of human verification, which makes their evaluation results less trustworthy. In contrast, \ourdataset is human-verified and features longer, detailed prompts that seek more specific information, thus posing greater challenges to LLMs by requiring them to reason across long-tailed facts.

We quantify these observations by breaking down our longer prompts into shorter, generic prompts that request basic details about the proper nouns involved. Ideally, LLMs equipped with all the necessary knowledge to solve prompts in our dataset should achieve perfect results on these decomposed prompts. Interestingly, we find that even the basic details of the proper nouns pose a challenge for LLMs, indicating that they also lack the necessary knowledge to solve our benchmark.

%% file: sections/related.tex
\section{Related Work}

Factuality evaluation has traditionally been focused on short-form question answering \citep{joshi-etal-2017-triviaqa,kwiatkowski-etal-2019-natural,lin-etal-2022-truthfulqa,vu2023freshllms,wei2024measuring}. Recently, there has been increasing interest in long-form factuality tasks. \citet{NEURIPS2022_df438caa} adapted the FEVER fact verification dataset \citep{thorne-etal-2018-fever} into prompts, prompting LLMs to generate continuations. They evaluated the factuality of these continuations based on the presence of named entities. \citet{manakul-etal-2023-selfcheckgpt} used templates to form prompts, asking LLMs to generate articles for Wikipedia entities and checking factuality through LLMs as well. In a similar setup, \citet{min-etal-2023-factscore} proposed using external knowledge sources during evaluation, though their focus was primarily on Wikipedia biographies. Recent long-form factuality benchmarks have begun exploring more diverse topics. \citet{bang2025hallulens} and \citet{ravichander2025halogen} compiled a benchmark from existing datasets and automatically created prompts to investigate different types of hallucinations. Unlike most prompt sets used in prior work, \ourdataset is human annotated and contains more diverse topics.

Perhaps the most related works to ours are \longfact \citep{NEURIPS2024_937ae0e8}, where LLMs are prompted to generate questions based on a list of predefined topics, and \factbench \citep{bayat2024factbench}, where existing datasets are automatically filtered to construct the dataset. Unlike these works, which contain only automatically constructed prompts and where state-of-the-art LLMs can already achieve saturated performance on most prompts, \ourdataset is a human-verified, challenging prompt set.

%% file: sections/factory.tex
\section{\ourdataset}

\begin{figure}
    \centering
    \includegraphics[width=\linewidth]{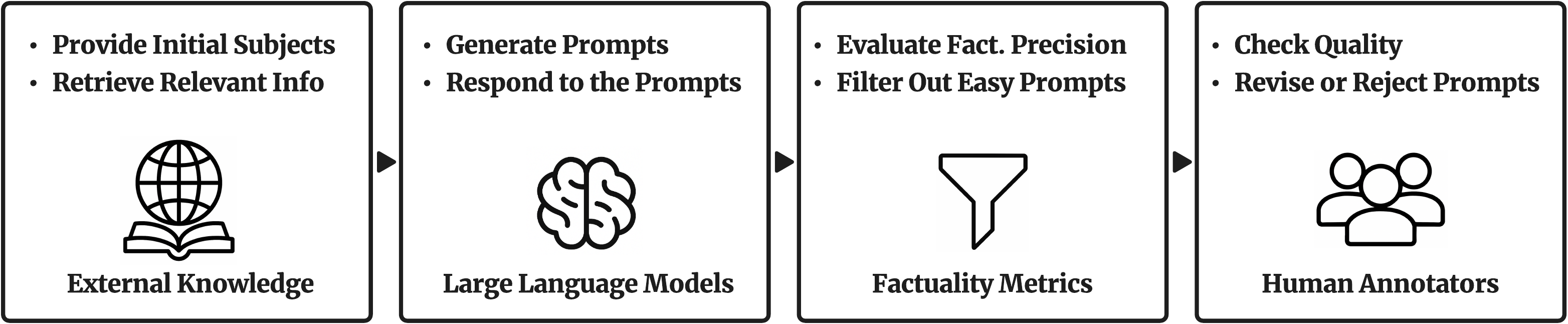}
    \caption{Diagram illustrating our data generation pipeline. To tackle the challenges of long-form evaluation, we employ a model-in-the-loop approach, enabling rapid and comprehensive assessments of prompt quality.}
    \label{fig:dataset_gen_pipeline}
\end{figure}

\begin{figure}
    \centering
    \includegraphics[width=0.5\linewidth]{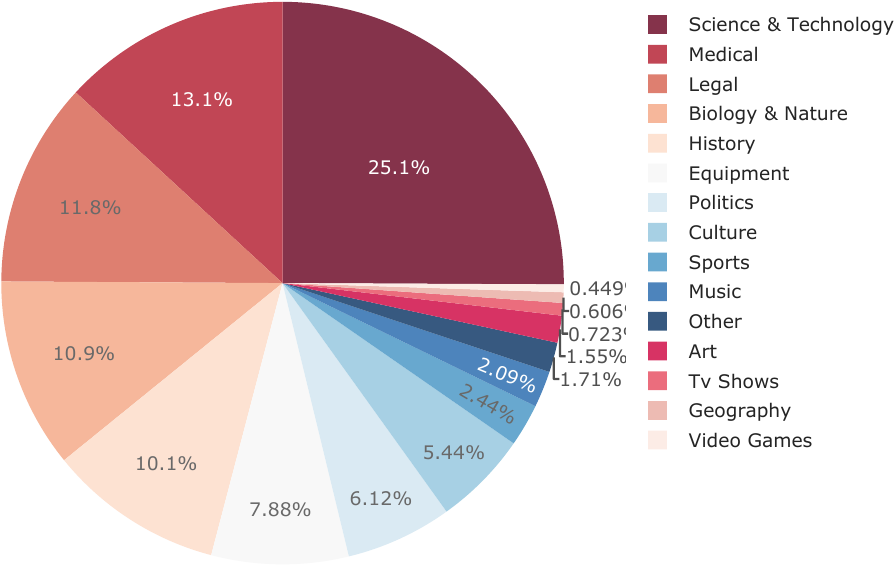}
    \caption{Distributions of prompt categories in \ourdataset.}
    \label{fig:dataset_pie_chart}
\end{figure}

One method for curating a set of diverse and challenging prompts is to have human editors directly interact with the model and revise the prompts iteratively until the model generates an incorrect response. 
While effective for short-form QA tasks like SimpleQA~\citep{wei2024measuring}, this method is impractical for prompts requiring long, factual responses due to the laborious process of verifying each claim.
As a result, we adopt a model-in-the-loop approach~\citep{nie-etal-2020-adversarial, kiela-etal-2021-dynabench} and create a pipeline to automatically identify prompts that are challenging to state-of-the-art models and then have human editors to verify and revise those candidate prompts. We primarily utilize Llama models\footnote{Llama-3.1-70B and Llama-4-Maverick-Instruct} throughout the pipeline, unless otherwise specified.

As shown in Figure~\ref{fig:dataset_gen_pipeline}, our prompt construction process, at a high level, involves generating candidate questions, automatically filtering out less challenging ones, and then having humans revise them to ensure quality. While the filtering and revising steps could be repeated multiple times for improved quality, we perform them only once as the resulting benchmark is already challenging for state-of-the-art LLMs to solve. To create a diverse set of prompts, it is essential to curate a varied collection of seed prompts before they undergo human revision. We begin with Wikipedia because it covers a broad range of topics and provides publicly available knowledge on the subjects. Specifically, we use the entire set of Wikipedia titles\footnote{We use the 2021 Wikipedia dump from \citet{JMLR:v24:23-0037}, which consists of approximately 33 million entries.} as queries to retrieve relevant information from \mds~\citep{NEURIPS2024_a5d8aba2}, and then prompt LLMs to generate questions. With these generated questions, we first use LLMs to respond to them. Prompts that LLMs find unanswerable, given the retrieved passages from \mds, are filtered out at this stage. We then employ \vs~\citep{song-etal-2024-veriscore} to fact-check each model's response,\footnote{The standard version of \vs relies on online search engines, which can be costly. To enable scalable usage, we adapt the prompting versions of the claim verifier from \vs for use with MassiveDS~\citep{NEURIPS2024_a5d8aba2}, while still using its finetuned claim extractors.} filtering out questions that LLMs can answer with high factual precision.\footnote{We retain prompts for which the Llama models achieve less than 60\% factual precision.} This approach ensures that our dataset remains challenging to current state-of-the-art LLMs. The remaining questions are then sent to human annotators for revision.

\begin{table}[t]
    \centering\small
    \begin{tabular}{c|c|c|c|c}\toprule
         &  Human Verified & Scale & Difficulty \\\midrule
     \longfact~\citep{NEURIPS2024_937ae0e8}   & \xmark & 2k & Easy  \\
     \factbench~\citep{bayat2024factbench} & \xmark & 1k & Medium \\ 
     \ourdataset & \cmark & 10k & Hard  \\ \bottomrule
    \end{tabular}
    \caption{Comparison between \ourdataset and previous long-form factuality benchmarks. This comparison highlights that \ourdataset is large-scale, human-verified, and presents a greater challenge than earlier datasets.}
    \label{tab:dataset_comparison}
\end{table}

For human annotations, annotators are instructed to make necessary edits or even reject prompts to ensure that each question meets the following criteria: (a) it is fact-seeking (rather than, for example, asking hypothetical questions or seeking help with creative writing); (b) it is self-contained and does not need additional context to disambiguate the question; (c) it can be answered using public, trustworthy online information; (d) expected responses should not vary over time; (e) it does not seek to elicit unsafe responses. Additionally, annotators are instructed to provide URLs where they can find minimal relevant information to answer the prompts.\footnote{Although annotators are not expected to find information that exhaustively covers every aspect of the prompt, this exercise has proven helpful in improving annotation quality.} Annotators are instructed to reject prompts if a complete rewrite is necessary during the revision process.

A total of 39 human annotators participated in the annotation process, spending approximately 5 minutes on each prompt revision task. Ultimately, around 20\% of the prompts were rejected or edited according to different error categories, leaving us 10,156 prompts. To demonstrate the topic diversity of these prompts, we present the distribution of topic categories in Figure~\ref{fig:dataset_pie_chart} (see Appendix~\ref{sec:app:prompts} for details on the prompt used for topic classification). Additionally, the distribution of the types of human edits performed is provided in Appendix~\ref{sec:app:human_edit}.

To identify the most challenging subset of prompts, we apply further filtering to select those where a subset of our benchmarked state-of-the-art LLMs\footnote{We used Claude and GPT models.} achieves around 50\% factual precision. This subset, referred to as the ``hard'' split, consists of 421 prompts.

%% file: sections/experiments.tex
\section{Experiments}

\subsection{Experimental Setup}
We benchmark 6 state-of-the-art LLMs, including \claude\footnote{claude-3-7-sonnet-20250219}, \gemini\footnote{gemini-2.5-pro-preview-05-06}, \deepseek\footnote{https://huggingface.co/deepseek-ai/DeepSeek-V3-0324}, \gpt\footnote{gpt-4o-2024-06-01}, Qwen3\footnote{https://huggingface.co/Qwen/Qwen3-235B-A22B} and \llama\footnote{https://huggingface.co/meta-llama/Llama-4-Maverick-17B-128E-Instruct-FP8} on \longfact~\citep{NEURIPS2024_937ae0e8}, \factbench~\citep{bayat2024factbench} and \ourdataset.
Additionally, we apply retrieval augmentation to these models by retrieving the top 20 most relevant passages from \mds,\footnote{We use 9 domains, namely: DPR Wiki~\citep{karpukhin-etal-2020-dense}, Math~\citep{welleck2021naturalproofs,paster2023openwebmath}, Pes2o~\citep{lo-etal-2020-s2orc,peS2o}, PubMed~\citep{pubmed}, RPJ Books~\citep{NEURIPS2024_d3449733}, RPJ arXiv~\citep{NEURIPS2024_d3449733}, RPJ C4~\citep{NEURIPS2024_d3449733}, RPJ GitHub~\citep{NEURIPS2024_d3449733}, RPJ StackExchange~\citep{NEURIPS2024_d3449733}.} using Contriever~\citep{izacard2022unsupervised} as the embedding model and input prompts as the queries. For all models and datasets, we use the prompt ``\emph{In the response, provide as many specific details and examples as possible (such as names of people, numbers, events, locations, dates, times, etc.)}'' and generate up to a maximum of 1,024 steps, similar to \citet{chen2024improving}. For \longfact, we use the subtset of 250 prompts provided by \citet{NEURIPS2024_937ae0e8}. For \factbench, our preliminary experiments indicate that SOTA LLMs have achieved saturated performance (over 90\% precision) on the non-hard splits. Therefore, we only report model performance on the hard split, which consists of 532 prompts.

We use human evaluations to verify the factual precision for the benchmarks. We do so by randomly sampling 100 sentences from each model's output. These sentences, along with the automatically extracted claims, are presented to human annotators. Annotators have the ability to edit the claims before making their judgments, ensuring that the claims accurately reflect the content of the sentences. They then review and judge the claims (using online search engines) according to the same three labels used in \vs (``supported,'' ``unsupported'' or ``inconclusive''). This annotation task employed 25 annotators and each sentence took approximately 12 minutes.

\subsection{Benchmarking Results}

\begin{figure}
    \centering
    \includegraphics[width=\linewidth]{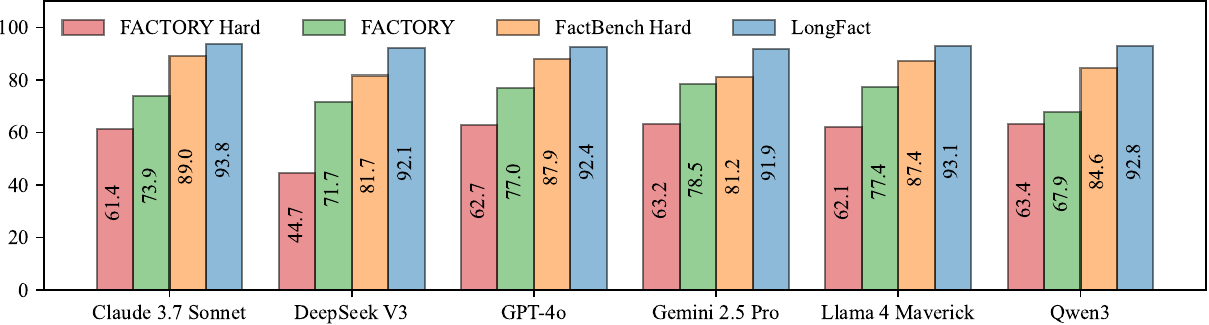}
    \caption{Factual precision as evaluated by human annotators on 100 sentences per model for each benchmark. All the models are retrieval-augmented.}
    \label{fig:benchmark-performance}
\end{figure}

We present benchmarking results in Figure~\ref{fig:benchmark-performance}. As expected, most state-of-the-art LLMs achieve over 90\% factual precision on \longfact, suggesting that their performance is approaching saturation. Additionally, \factbench Hard slightly reduces the performance of LLMs to approximately 85\%.
It is important to note that since \factbench and \longfact are automatically constructed, the lower factual precision could partly be due to the limited quality of the prompt set (see Section~\ref{sec:analysis} for more discussions on the quality issues). In contrast, \ourdataset presents a greater challenge than both \factbench Hard and \longfact, further decreasing LLM performance to about 75\%, while also being much larger in scale. With \ourdataset Hard, the performance of state-of-the-art LLMs drops to around 60\%. In addition, we do not observe significant differences in the length of model outputs across various benchmarks (see Appendix~\ref{sec:app:model_output_lengths} for more details), suggesting that the lower factual precision on \ourdataset does not come from longer model generations.\footnote{Suppose responses of the same length make similar number of claims and the number of facts to be included about each prompt is a constant. A lower factual precision indicates a lower recall when the length of the response remains the same.}

\subsection{Analysis}\label{sec:analysis}

\paragraph{What makes \ourdataset different from \longfact and \factbench?}

\begin{figure}[t]
    \centering
    \begin{tabular}{p{.15\linewidth}|p{.6\linewidth}|p{.15\linewidth}}\toprule
    Dataset & Prompt & Issue \\ \midrule
       \longfact  & 1. Who is Emilia Chico? & Not Answerable \\
       & 2. Can you provide an overview of the International Monetary Fund? & Not Challenging \\ 
       & 3. What was the role of the United States in the Treaty of Tordesillas? & Hallucination \\\midrule
       \factbench Hard  & 1. What are the strongest countries in Civilization 5? & Subjective \\
        & 2. The latest 50 kernel versions and release times of Linux & Time Sensitive \\
       & 3. What might happen as a consequence if the VAT on public transport services was reduced e.g. from 25\% to 10\%? & Hyperthetical \\
\bottomrule
\end{tabular}
\vspace{3em}
\begin{minipage}{.7\textwidth}
\begin{tabular}{p{\linewidth}}
\toprule
     Explain the legal framework established by the Protection from Eviction Act 1977 in the United Kingdom regarding tenant rights. \\\midrule
     What challenges did Benni McCarthy encounter during his initial seasons as head coach of Cape Town City FC, and what were the team's performances in league and cup competitions? \\
     \bottomrule
\end{tabular}
\end{minipage}\hfill\begin{minipage}{.3\textwidth}
\hfill\includegraphics[width=0.9\linewidth]{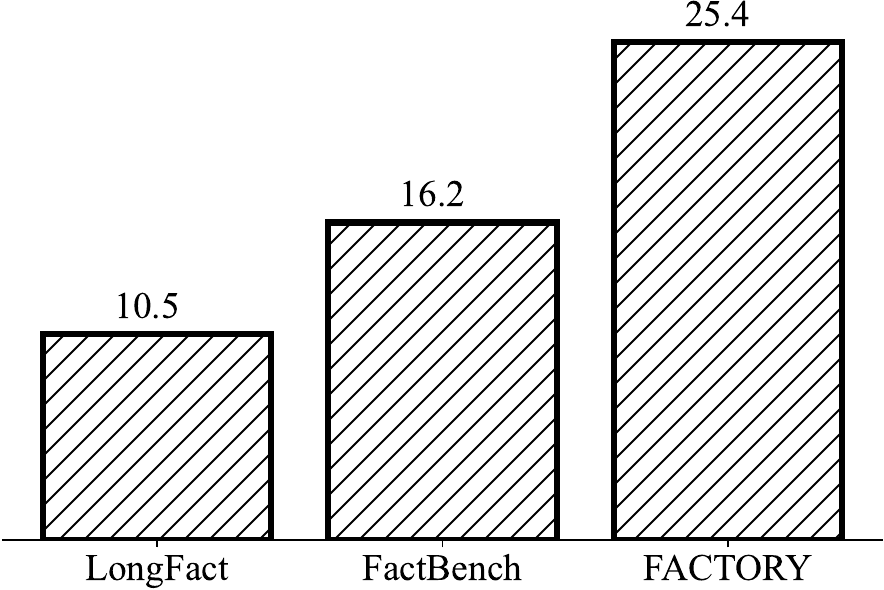}
\end{minipage}\vspace{-3.5em}
    \caption{Top: Examples illustrating potential quality issues of existing factuality benchmarks. Bottom Right: Average prompt lengths. Compared to \longfact and \factbench, \ourdataset has significantly longer and more detailed prompts. Bottom Left: Example prompts from \ourdataset.}
    \label{fig:dataset_quality_examples}
\end{figure}

We explore specific examples that distinguish \ourdataset from other existing long-form factuality benchmarks. Firstly, we emphasize that \ourdataset addresses several issues found in previous benchmarks, such as answerability, difficulty, and the querying of time-sensitive information. Figure~\ref{fig:dataset_quality_examples} presents examples illustrating these issues. For instance, the first question from \longfact is unanswerable because ``Emilia Chico'' is not a well-known figure. Conversely, the second question is too easy, as there is a Wikipedia page about the International Monetary Fund. Retrieval-augmented LLMs could simply return the introductory paragraph from that page to provide an accurate response. The last question contains hallucinations. The phrase ``United States of America'' was coined in 1776, the same year the US declared its independence. Therefore, it could not have been involved in the Treaty of Tordesillas, which was signed in 1494.

Examining examples from \factbench, the first prompt seeks the LLMs' subjective opinion on the strongest countries in a game. While seeking subjective opinions can be acceptable if there is a public consensus, it rarely is the case and often leads to challenges in evaluating factuality, as finding supporting evidence can be difficult. With supporting information being absent, model outputs will be considered inconclusive, leading to lower factual precision without making the prompt set actually more challenging. Therefore, when constructing \ourdataset, we explicitly instruct human annotators to edit or reject prompts that seek subjective advice. To assess the prevalence of this issue, we calculated the fraction of prompts containing words related to subjective opinions, such as ``best'' and ``top.'' Our findings revealed that 16\% of \factbench's hard prompts seek subjective opinions. This suggests that the difficulty of these hard prompts may partly stem from their inherent subjectivity. The second example is time-sensitive, as it requests the latest information. The final example is hypothetical, making it unsuitable for evaluation under the factuality setting.

In contrast, as shown in Figure~\ref{fig:dataset_quality_examples}, prompts in \ourdataset are significantly longer, seek more specific details, and are thus more challenging to existing LLMs.

\paragraph{Why do LLMs fail on \ourdataset?}

\begin{figure}[t]
    \centering\includegraphics[width=.7\linewidth]{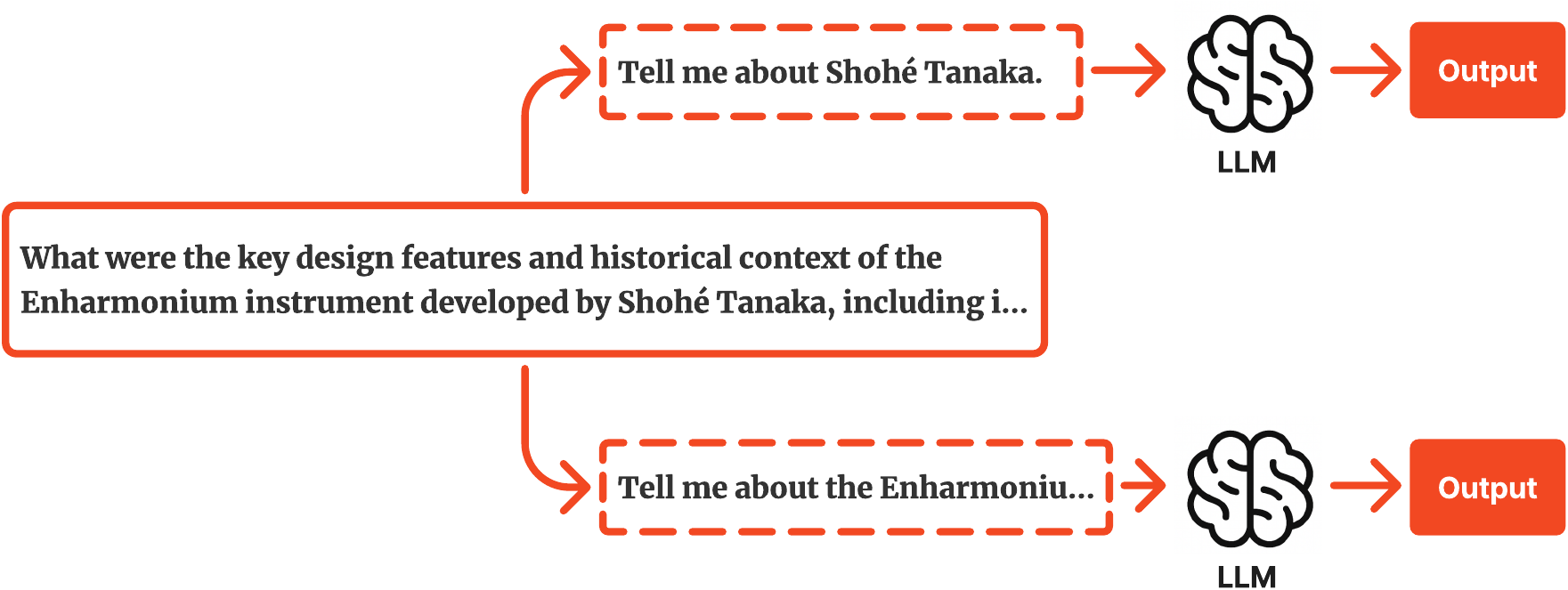}
    \caption{Diagram illustrating the evaluation pipeline of atomic prompts. The atomic prompts are automatically generated by instructing language models to build generic questions using proper nouns in the original prompts. These atomic prompts are then used independently to assess the language models' knowledge of specific subjects.}
    \label{fig:breakdown-examples}
\end{figure}

\begin{figure}[t]
    \centering
    \includegraphics[width=.3\linewidth]{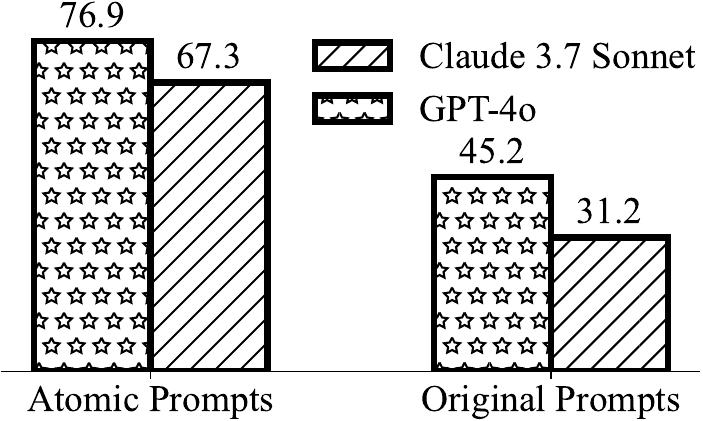}
    \caption{Comparison of factual precision when evaluated on atomic prompts versus original prompts from the \ourdataset Hard dataset.}
    \label{fig:metric_results}
\end{figure}

To determine whether the challenges in \ourdataset arise from the long, detailed style of prompts (which necessitates reasoning capabilities across different facts) or the long-tailed knowledge required, we conducted additional experiments. We transformed the detailed prompts into more generic ones for each proper noun mentioned in the prompt, creating what we call ``atomic prompts,'' similar to the question styles in \longfact and \factbench.
Ideally, LLMs equipped with all the necessary knowledge to solve prompts in our dataset should achieve perfect results on these atomic prompts.
The evaluation process is demonstrated in Figure~\ref{fig:breakdown-examples}.  This process was applied to \ourdataset Hard, and we randomly sampled 100 prompts for evaluation using \vs~\citep{song-etal-2024-veriscore}.\footnote{When using \vs, we employ their finetuned claim extractor in combination with the \gpt claim verifier, as this setup performs better than other combinations.} On average, each prompt was broken down into two prompts. We evaluated \gpt and \claude on these prompts without using RAG and showed the results in Figure~\ref{fig:metric_results}. Interestingly, while making the prompts more atomic helps in making them more solvable, they remain relatively challenging compared to current long-form factuality benchmarks. Notably, factual precision of the responses to \ourdataset prompts is substantially lower as more specific information is needed. Such prompts requires models to reason across various facts, thereby increasing the difficulty of the task. These findings suggest that \ourdataset is challenging to solve due to both its long-tailed knowledge and the reasoning capabilities required.

%% file: sections/conclusion.tex
\section{Conclusion}

In this work, we addressed the limitations of existing long-form factuality benchmarks by introducing \ourdataset, a human-verified prompt set designed to challenge state-of-the-art LLMs. Our findings demonstrate that \ourdataset significantly reduces factual precision compared to existing benchmarks, underscoring its difficulty and the need for improved model capabilities. Our analysis highlights the importance of human verification in benchmark development and identifies key issues in prior benchmarks, such as answerability and hallucinations. By requiring models to reason across detailed, specific information, \ourdataset sets a new standard for evaluating long-form factuality, paving the way for future advancements in model development and evaluation.

%% file: sections/appendix.tex
\section{Prompts}\label{sec:app:prompts}

\begin{figure}[h]
\begin{tcolorbox}[colback=black!5!white,colframe=black!75!black,title=Prompt for Generating Initial Questions]
\begin{VerbatimWrap}
Instructions:
1. Given a list of documents, you are tasked to ask a few generic, fact-seeking question about "{object}" that are challenging yet still answerable.
2. Ensure the question prompts long-form responses of approximately 1024 tokens and avoiding seeking subjective opinions or predictions about future trends.
3. Avoid asking time-sensitive questions; instead of using terms like "now", "today", or "current," specify the exact date. Dates are not needed for events that have already occurred.
4. Eliminate any ambiguity in the question, especially regarding the names of events or individuals, as multiple entities may share the same name.
5. Make sure the question is self-contained and does not reference the text provided in these instructions and the relevant documents, or in prior generated questions.
6. Avoid asking "why" and "how" questions.
7. Avoid asking questions that require knowledge beyond 2019.
8. Avoid including too many details in the question as it may leaks relevant information.
9. Instead of providing the exact date, simply mention the year an event took place.
10. Try your best to make sure that the questions can be answered using the documents, and it's better to draw information from multiple documents instead of depending on just one. However, if the provided documents lack relevant information, simply reply with "No relevant information!".
11. Avoid asking questions about the impact, influence, implication, or significance of certain events; instead, ask questions like those seeking details of an event, the features of an object, or the biography of a person.
12. Please prepend "Question: " to the start of each question that is asked.

EXAMPLES:
Question: What is the Stanford Prison Experiment?
Question: Explain the use of the Atomic Bomb on during World War II.
Question: What is the Facebook's data privacy scandal in 2018?
Question: Tell me about the Thalidomide drug scandal in the 1950s and 1960s.
Question: What unfolded during the Oklahoma City bombing in 1995?

The examples above are provided to help you understand the high-level requirements of the task. You do not need to reuse the exact phrases or follow the same style of the example questions.

Documents:

{docs}
\end{VerbatimWrap}
\end{tcolorbox}
\label{app:fig:prompt-question-gen}
\end{figure}

\begin{figure}[h]
\begin{tcolorbox}[colback=black!5!white,colframe=black!75!black,title=Prompt for Generating Responses]
\begin{VerbatimWrap}
Instructions:
1. Please respond to the following question.
2. In the response, provide as many specific details and examples as possible (such as names of people, numbers, events, locations, dates, times, etc.)

Question: {question}
\end{VerbatimWrap}
\end{tcolorbox}
\label{app:fig:prompt-respond}
\end{figure}

\begin{figure}[h]
\begin{tcolorbox}[colback=black!5!white,colframe=black!75!black,title=Prompt for Generating Retrieval-Augmented Responses]
\begin{VerbatimWrap}
Instructions:
1. Please respond to the following question based on the provided documents.
2. In the response, provide as many specific details and examples as possible (such as names of people, numbers, events, locations, dates, times, etc.)

Question: {question}

Relevant Documents:

{docs}
\end{VerbatimWrap}
\end{tcolorbox}
\label{app:fig:prompt-rag-respond}
\end{figure}

\begin{figure}[h]
\begin{tcolorbox}[colback=black!5!white,colframe=black!75!black,title=Prompt for Categorizing Text]
\begin{VerbatimWrap}
Given a short text, determine the single most appropriate category from the following options: TV Shows, Music, Sports, Geography, Culture, Art, Politics, Science & Technology, Video games, History, Medical, Legal, Equipment, Biology & Nature, and Other. When generating the category, use a new line and append "Category: " to the begining of the line. Below are the defitions of each category and the text.

Category Definition:

- TV Shows: Assign prompts related to television series, including specific shows, characters, or networks.
- Music: Use this category for prompts about songs, artists, albums, genres, or music events.
- Sports: Categorize prompts that involve athletic activities, teams, players, or sporting events.
- Geography: This category is for prompts about locations, countries, cities, landmarks, or geographical features.
- Art: Assign prompts related to visual arts, including painting, sculpture, photography, or art movements.
- Politics: This category should cover political ideologies, political parties, elections, governance, policy-making, and political figures. It involves the processes and activities associated with government and decision-making at various levels.
- Legal: This category should focus on the judicial system, laws, legal processes, court cases, and the roles of legal professionals. It deals with the application and interpretation of laws and legal principles.
- Science & Technology: Categorize prompts that involve scientific concepts, discoveries, technological advancements, or related figures.
- Video Games: This category is for prompts about video games, gaming consoles, developers, or gaming culture.
- History: Assign prompts related to historical events, figures, periods, or artifacts.
- Medical: Use this category for prompts about health, medicine, diseases, treatments, or medical professionals.
- Equipment: This category includes devices and machinery used in various applications, such as cameras, vehicles, consumer electronics, industrial machinery, and tools.
- Culture: This cateogry encompasses customs, traditions, social behaviors, and shared values of different groups and societies, including cultural practices, languages, cuisine, fashion, and cultural heritage.
- Biology & Nature: This category covers prompts related to living organisms, their environments, and ecological interactions. It includes topics on species, ecosystems, conservation, and biodiversity.
- Other: Use this category for prompts that do not fit into any of the specified categories.

Text:

{text}
\end{VerbatimWrap}
\end{tcolorbox}
\label{app:fig:prompt-categorize-text}
\end{figure}

\begin{figure}
\begin{tcolorbox}[colback=black!5!white,colframe=black!75!black,title=Prompt for Converting Prompts into Atomic Prompts]
\begin{VerbatimWrap}
Please extract the proper nouns from the following question and create questions that inquire about each specific proper noun. The questions should be concise and general, such as "Tell me about Computer Science." Ensure that you ask exactly one question for each proper noun found in the original question. The questions should be clear and comprehensible on their own, without any ambiguity. Begin each question with "Question: ".

"{question}"
\end{VerbatimWrap}
\end{tcolorbox}
\label{app:fig:prompt-breakdown}
\end{figure}

\section{Model Output Lengths}\label{sec:app:model_output_lengths}

\begin{figure}[h]
    \centering
    \includegraphics[width=\linewidth]{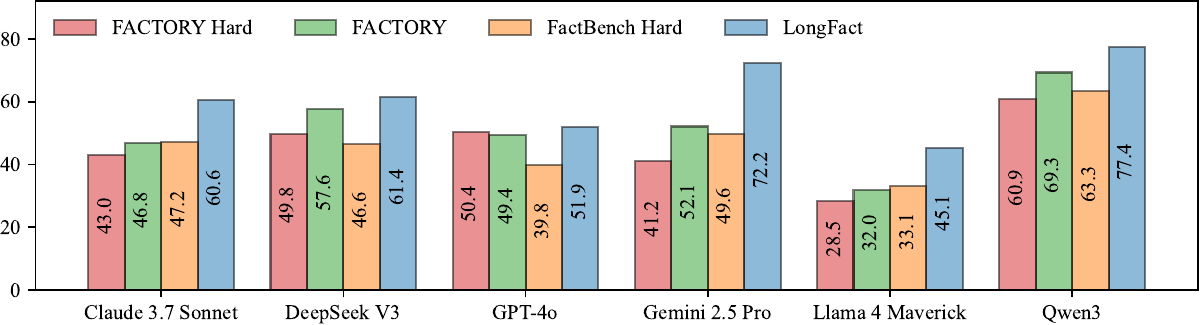}
    \caption{Average numbers of claims (produced by \vs) for model outputs.}
    \label{fig:app:num_claims}
\end{figure}

\begin{figure}[h]
    \centering
    \includegraphics[width=\linewidth]{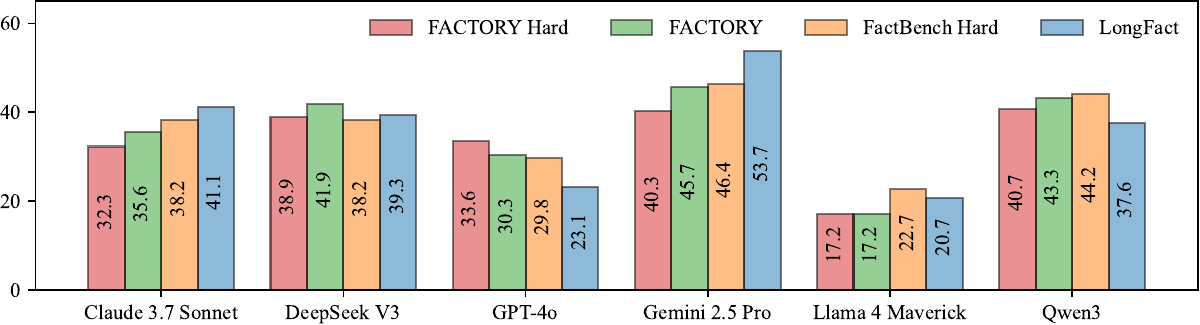}
    \caption{Average numbers of sentences for model outputs.}
    \label{fig:app:num_sents}
\end{figure}

We report the average number of sentences and the average number of claims (produced by \vs) in model outputs across different benchmarks, shown in Figure~\ref{fig:app:num_claims} and Figure~\ref{fig:app:num_sents}, respectively. Overall, except for \longfact, models generate outputs of similar length across the various benchmarks.


\section{Human Edits in Prompt Annotations}\label{sec:app:human_edit}

\begin{figure}
    \centering\includegraphics[width=0.5\linewidth]{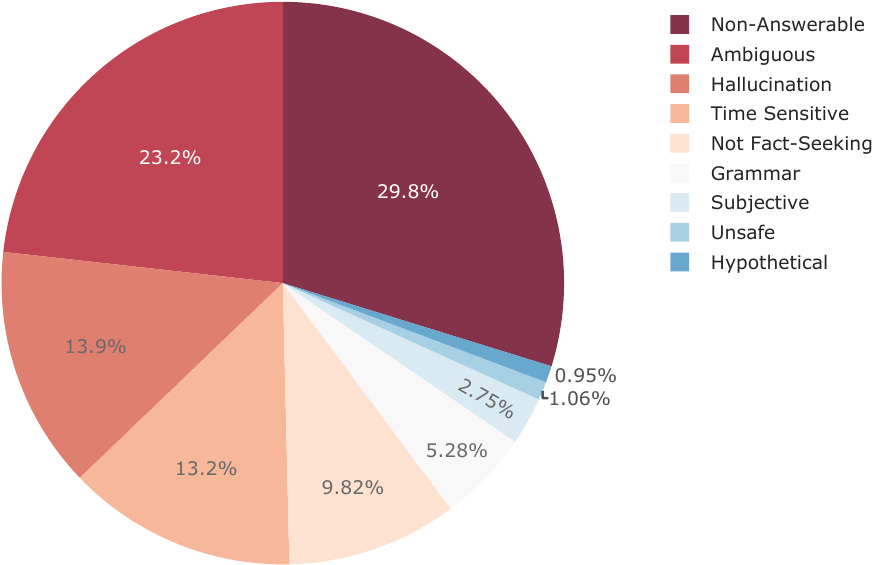}
    \caption{Distributions of the types of human edits.}
    \label{fig:human_edits_pie_chart}
\end{figure}

We present the distribution of human edit types made during prompt annotation in Figure~\ref{fig:human_edits_pie_chart}.